\documentclass[letterpaper, 10 pt, conference]{ieeeconf} 
\IEEEoverridecommandlockouts
\usepackage{amssymb}
\usepackage{amsmath}  
\usepackage{comment}
\usepackage{fancyhdr}
\usepackage{tikz}
\usepackage[utf8]{inputenc}
\usepackage[T1]{fontenc}
\usepackage{comment}
\usepackage{graphicx} 
\usepackage{dblfloatfix} 
\graphicspath{ {Figures/} }
\usepackage{amsmath, amsfonts, bm}
\usepackage{xcolor}
\usepackage{array}
\usepackage{setspace}
\newcolumntype{L}{>{\centering\arraybackslash}m{0.45in}}
\newcolumntype{W}{>{\centering\arraybackslash}m{0.8in}}
\usepackage{threeparttable}
\usepackage{hyperref}
\usepackage{float}
\usepackage{caption}
\usepackage{subcaption}
\usepackage{accents}
\usepackage{cite}
\usepackage{xurl}

\DeclareSymbolFont{symbolsC}{U}{txsyc}{m}{n}
\DeclareMathSymbol{\circleleft}{\mathrel}{symbolsC}{146}
\DeclareMathSymbol{\circleright}{\mathrel}{symbolsC}{145}

\title{\LARGE \bf Characterization and Correlation of Robotic Snake Scale Friction and Locomotion Speed}

\author{\"Umit \c{S}en$^{1}$, Andri Mahegan$^{1}$, and Gina Olson$^{1}$%
\thanks{$^{1}$Department of Mechanical and Industrial Engineering, University of Massachusetts, Amherst, MA, USA.}}

\begin{document}

\maketitle
\begin{tikzpicture}[remember picture,overlay]
\node[anchor=south, yshift=20pt] at (current page.south) {
    \parbox{\dimexpr\textwidth-2cm\relax}{
        \centering\footnotesize \copyright~2026 IEEE. Personal use of this material is permitted. Permission from IEEE must be obtained for all other uses, in any current or future media, including reprinting/republishing this material for advertising or promotional purposes, creating new collective works, for resale or redistribution to servers or lists, or reuse of any copyrighted component of this work in other works.
    }
};
\end{tikzpicture}

\thispagestyle{empty}
\pagestyle{empty}

\begin{abstract}

Snake robots are inspired by the ability of biological snakes to move over rock, grass, leaves, soil, up trees, along pavement and more.
Their ability to move in multiple distinct environments is due to their legless locomotion strategy, which combines distinct gaits with a skin that exhibits frictional anisotropy. 
Designing soft robotic snakes with similar capabilities requires an understanding of how this underlying frictional anisotropy should be created in engineered systems, and how variances in the frictional anisotropy ratio affect locomotion speed and direction on different surfaces.
While forward and backward frictional ratios have been characterized for previous scale designs, lateral friction and the associated ratios are often overlooked.
In this paper, our contributions include: 
(i) the development of a novel articulated pseudo-skin design that is modular, easy to construct and has removable or replaceable scales;
(ii) experimental measurement of the frictional characteristics of otherwise-identical scales at varying angles of attack (15°, 25°, 35°, 45°) on different surfaces of interest (grass, bark, smooth surface, carpet);
(iii) separate measurements of locomotion speed for each angle and surface.
Consequently, while we observed some consistent trends between frictional coefficients and scale angle, aligning with literature and intuition, we were not able to consistently identify expected correlations between frictional ratios and locomotion speed. 
We conclude that either frictional ratios alone are not sufficient to predict the observed speed of a snake robot, or that specific measurement approaches are required to accurately capture these ratios. 

\end{abstract}


\section{INTRODUCTION}

Snakes are an inspiration for robots due to their ability to navigate vastly different environments, including rocky mountains and caves, forest floors, plant canopies, grass, etc, which is due in large part to their compliance \cite{marvi_sidewinding_2014,bridging_tree_snakes,SnakeOverall24,byrnes_gripping_2014}.
Snake locomotion relies on a combination of body movement patterns, or gaits, and anisotropic friction between the snake's belly and the ground, which is created by the belly scales \cite{hu_mechanics_2009}. 
Studies of snakes have shown that these belly scales exhibit different frictional properties in the forward, backward and lateral directions, and the evidence from a dynamic model suggests that snakes utilize this anisotropy to propel themselves \cite{hu_mechanics_2009}.

Snake-inspired soft robots have mimicked many aspects of snake biology, including backbones, actuation and scales, and have demonstrated lateral undulation, rectilinear and sidewinding gaits \cite{parvaresh_metamaterial_2024,branyan_snake-inspired_2020,marvi_sidewinding_2014,ta_design_2018,lamping_frictional_2022}. 
However, these robots continue to drastically under perform compared to their biological counterparts. 
While work is needed in many areas, scales are a critical and understudied component. 
Many works demonstrate frictional anisotropy between the forward and backward directions, but lateral friction is rarely measured, even though live-snake experiments supported by dynamic models \cite{hu_mechanics_2009} identify frictional anisotropy between the lateral and forward directions as an important predictor for snake locomotion performance. The literature also lacks an in-depth, empirical examination of how variances in frictional coefficients affect practical locomotion. 
In this work, we develop a new modular skin-and-scale design and use it to characterize friction and locomotion speed across scale angles and surfaces. 
We evaluate observed correlations, as a step to understanding how to formally design scales.

\begin{figure}[!t]
    \includegraphics[width=.95\columnwidth]{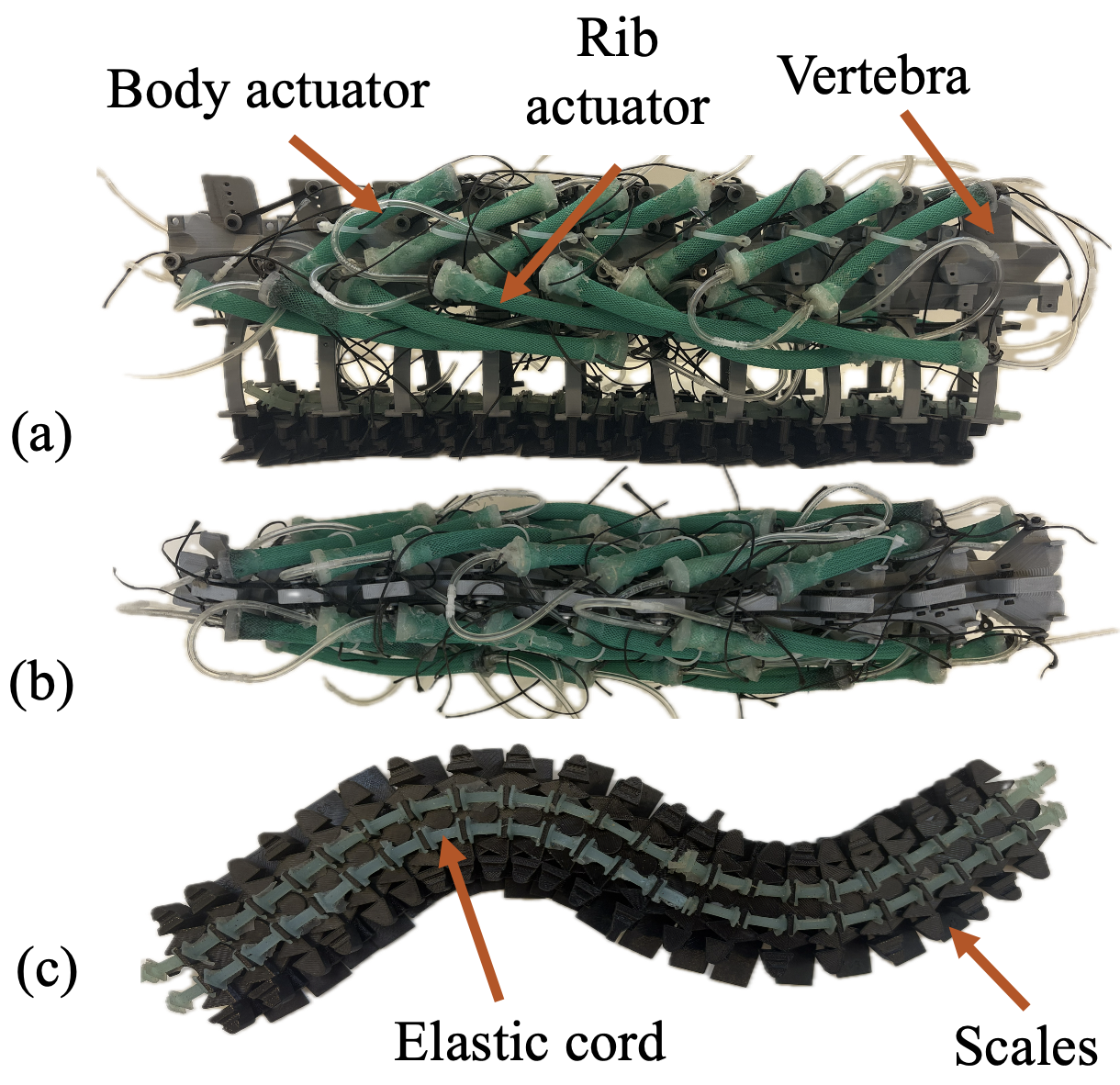}
    \caption{The snake robot ((a) top view; (b) side view; (c) scales) used herein has an articulated backbone and is actuated by air-driven McKibbens. The modular scale system is attached to the ribs extending from each vertebra.}
    \label{fig:frontSideTop}
    \vspace{-1.5em}
\end{figure}

\vspace{0.5em}
\noindent \textit{Related Works:}
Early studies on snake locomotion proposed that snakes mainly use their body to create push points against environmental protrusions (e.g., rocks, branches) to propel themselves forward \cite{gray_mechanism_1946}.
However, this theory was not able to explain how snakes moved across terrain without such protrusions (e.g., pavement, sand, dirt). 
More recent research has demonstrated that this type of locomotion can be explained by the frictional properties of snake scales \cite{hu_mechanics_2009}.
Hu et. al measured friction coefficients in the forward, backward and lateral directions for live, sedated snakes, and used a theoretical dynamic model to demonstrate that these frictions coefficients, when paired with empirical gait kinematics, result in forward movement.   
Studies since have observed that snakes adjust their scale angle to regulate anisotropy for enhanced locomotion \cite{marvi_snakes_2016} and that micro- and nano-scale features on the scales contribute to anisotropy \cite{wu_variation_2020}.

Soft snake robots have utilized different approaches to integrate frictional anisotropy.
We focus this review on scale-based designs, which are are ideologically similar to biological snakes, as opposed to those that use wheels to create anisotropy \cite{onal_luo_3D_snake}.
Artificial scales or scale-replacements include 3D-printed biomimetic scales \cite{chitikena_anisotropic_2024,huq_motion_2018}, kirigami-inspired scales \cite{rafsanjani_kirigami_2018,parvaresh_metamaterial_2024,branyan_snake-inspired_2020}, and dual-material approaches that use both a high friction and low friction material on the scale surface  \cite{ta_design_2018,chitikena_anisotropic_2024}.
Design and testing procedures, including which frictional values were measured and which gaits and surfaces were used, vary significantly across these works, which is expected given the early stage of the research. 
Parvaresh et al. measured the forward and backward frictional coefficients of a kirigami-created scale on foam surfaces.
Only one scale design was used, and robot speed - from a rectilinear gait - was correlated to environmentally-created differences in friction \cite{parvaresh_metamaterial_2024}. 
A prior work on kirigami scales for crawling examined frictional forces during movement in more depth, and noted a stick-slip behavior, complicating the process of measurement, and further found that distance traveled was maximized when the frictional anisotropy ratio was at its highest \cite{rafsanjani_kirigami_2018}. 
Branyan et al. adapted kirigami scales for lateral undulation, and also measured forward and backward friction \cite{branyan_snake-inspired_2020}.
Ta et al. measured forward, backward and lateral friction coefficients on three different surfaces (fabric, paper and white board) for their multi-material scales and reported robot speed.
However, as the work focused on comparing designs, the authors did not correlate frictional ratios to locomotion speed.

While many artificial scales are designed to a set angle, a few works have investigated alterable or a range of scale angles \cite{shen_design_2021,lamping_frictional_2022}. 
Shen et al. used dorsal-style scales (as opposed to belly, or ventral), paired with a crawling gait, and altered scale angle using a pressurized airbag under the scales. Forward and backward friction tests were conducted on a sandpaper and asphalt, and while friction measurements varied only slightly with pressure until a sudden jump, speed and scale angle changed near linearly with pressure. Lateral friction was not measured \cite{shen_design_2021}.
Lamping et al. investigated the influence of scale angle using manually set angles on a single segment robot performing a rectilinear gait. These measurements were performed on denim, and lateral friction was again neglected. 
While altering scale angle created clear differences in friction for most designs tested in the work, the authors did not measure locomotion speed for all cases.

\vspace{0.5em}
\noindent \textit{This Work:}
While prior works explore numerous aspects of the concept that frictional variation combined with gait can create locomotion, attempts to correlate empirical friction values and ratios to measured locomotion speed are surprisingly rare, and attempts to do so in natural environments are almost non-existent. 
Evidence that frictional ratios are key determiners of locomotion speed is strong in biological systems \cite{hu_mechanics_2009}, but these studies were conducted in the context of known gaits and body designs validated by evolution.
In designing artificial snake robots, roboticists must determine what frictional ratios \textit{should be} for their body design and gait. 
Studies that experimentally confirm a strong relationship between these frictional ratios and locomotion speed, though, are rare, and in at least one case, have not confirmed any direct proportionality \cite{shen_design_2021}.
In this work, we attempt to more comprehensively address this question, as a first step to creating tools to formally design snake robot bodies and scales to meet locomotion speed requirements. 
We focus on a single body design (a 10-vertebra highly articulated soft snake robot, shown in Figure \ref{fig:frontSideTop}) and gait, with no closed loop control, and alter only scale angle. 
In order to support a measurement of friction across a wide range of scale angles - and, in future work, scale designs - we developed a modular skin-and-scale design that allows scales to be swapped without removing the skin from the robots.
This scale architecture allows easy replication and can easily be adapted for different snake robot designs.
Directional (forward, backward, lateral) friction measurements were conducted with scale angles ranging from $15^\circ$ to $45^\circ$ degrees on four different surfaces (bark, grass, smooth surface, carpet), and locomotion speed was measured with the same surfaces and scale angles (Section \ref{sec:experiments}).
While we found clear and consistent correlations between scale angle and friction ratios, these correlations were surface-dependent, and locomotion speed and friction ratios did not consistently correlate.
We discuss these results herein and provide a potential explanation and possible next steps (Section \ref{sec:discussion}; \ref{sec:conclusion}).


\section{SNAKE ROBOT DESIGN}
\label{sec:design}

\begin{figure*}[!t]
    \includegraphics[width=1\textwidth]{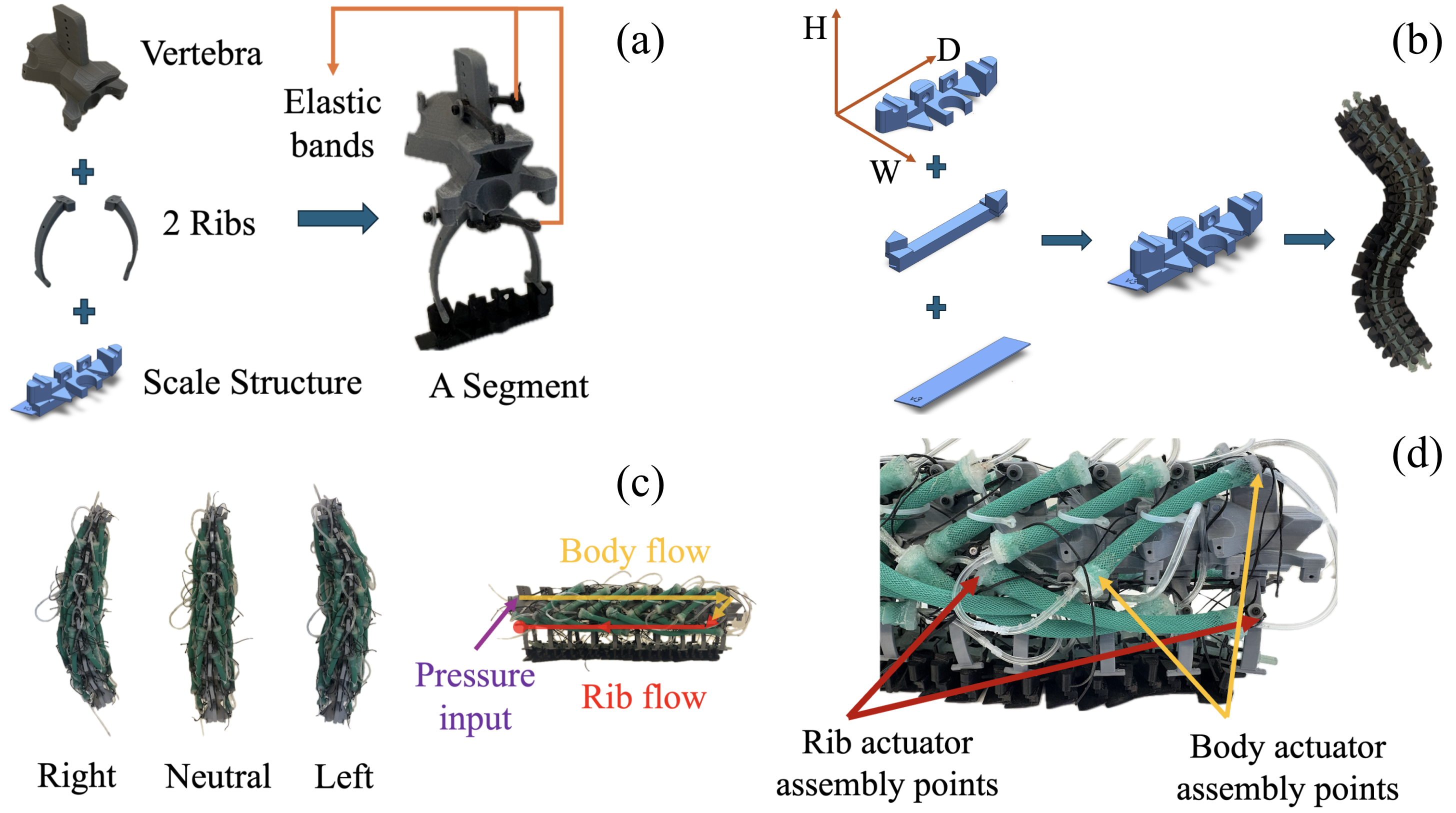}
    \caption{Vertebra, ribs, and scale structure forms one segment of the robot (a). Primary, secondary, and tertiary units assembled as one scale structure, and multiple scale structures connected through interlocking mechanism (b). 
    Right-side actuation, neutral (no-actuation), left-side actuation scenarios and air circulation scheme inside actuators (c). Assembly details (d). }
    \label{fig:combinedSnakeDesignAndActuation}
    \vspace{-1em}
\end{figure*}

Our bio-inspired snake robot has a modular design that consists of repeating vertebrae and rib structures \cite{previous_work}, and is more closely bio-inspired than most soft snake robots. McKibben actuators are used to mimic key muscle groups tied to locomotion, and we have added our modular skin-and-scale design to the base of the ribs. While the ideas build on that which was previously developed, the snake robot variant used herein is custom to this work. CAD files can be found in GitHub repository: \url{https://github.com/senumit/SnakeLocomotionDesignFiles/tree/main/SnakeRobot_DesignFiles}.
\subsection{Vertebrae and Ribs}

The vertebra design developed in this work mimics the proportions and features of arboreal colubrid snakes, such as \textit{bioga irregularis}, or brown tree snake. However, biological snake vertebra for these species are much smaller than those used in this paper, and have a linear density along the backbone that is - for the moment - beyond our manufacturing capabilities. The approximate size of our vertebra is 58 mm in height, 50 mm in width and 50 mm in depth (Figure \ref{fig:combinedSnakeDesignAndActuation}). 
Articulation occurs through a primary ball-and-socket joint, as well as additional articulation pairs common in snake robots: pre- and post-zygaphophysis and zygosphene/zygantrum.
The resulting motion is similar to two perpendicular revolute joints, or a stabilized universal joint: the relative angle between two vertebra can change in two perpendicular planes (in this design, by about 25$^\circ$ in either direction), but not at any angle in between, and allowable torsion is effectively negligible.
The bio-inspired design principles and mechanical architecture of the vertebrae are detailed in \cite{previous_work}.
While the vertebra permit body bending in either the horizontal or vertical plane, in this work we consider only bending in the horizontal plane, and our added scale design (described below) eliminates bending in the vertical plane. 
Two rib structures (approximately 41 mm in height, 7.5 mm in width, 22 mm in depth) are connected to each vertebral segment. 
The resulting design has 10 vertebral segments and 20 ribs as shown in Figure \ref{fig:combinedSnakeDesignAndActuation}(a). 
All vertebra and ribs were 3D printed using a PLA filament with a sparse infill density of 15\% and layer height of 0.2 mm.
Passive elastomeric bands molded from BBDino 30A are used to keep vertebra together. 

\subsection{Actuators}

\begin{figure}[!t]
    \includegraphics[width=1\columnwidth]{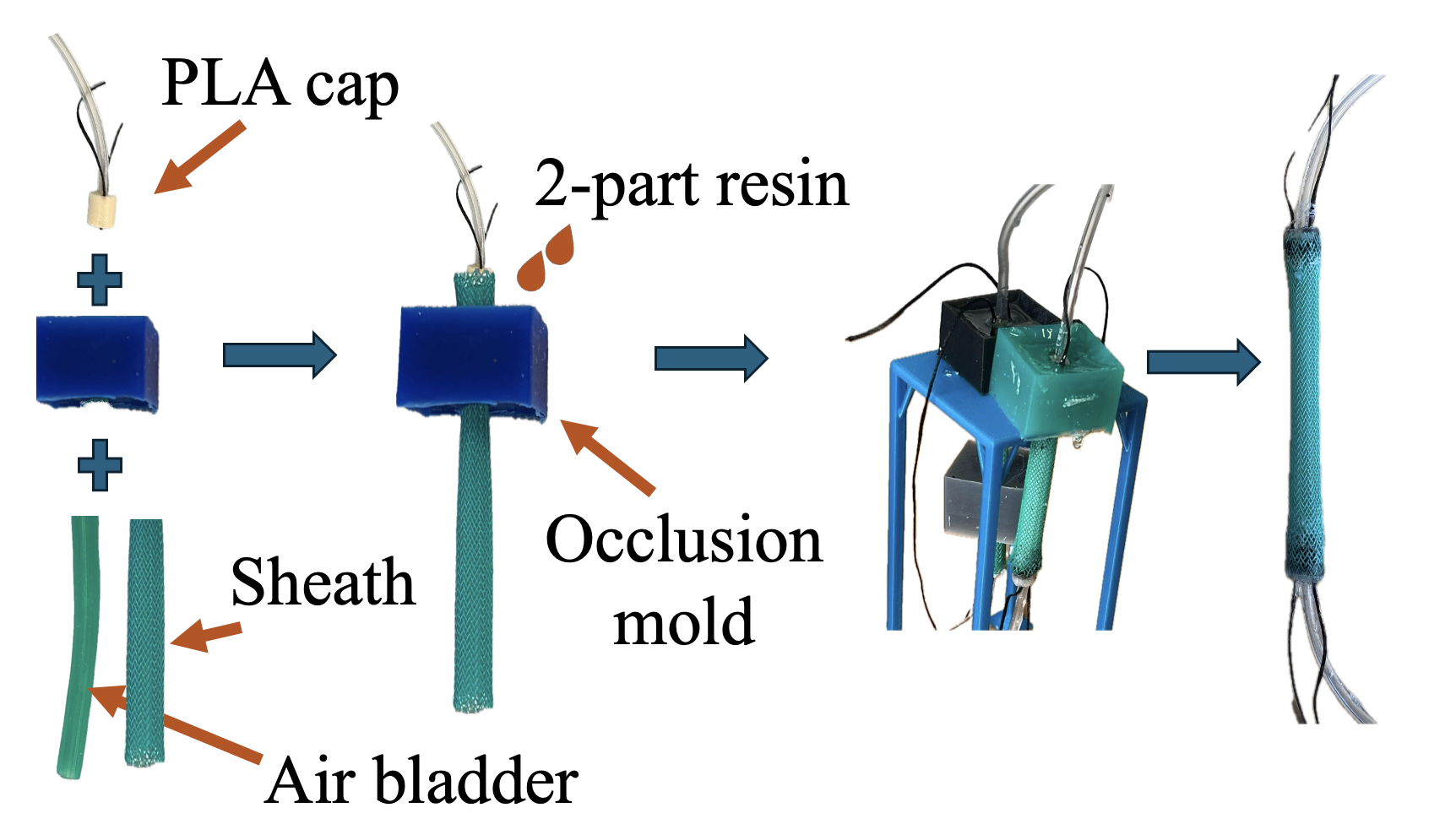}
    \caption{Capping procedure for McKibben actuators: the cap is bonded to the sheath using a 2-part resin, requiring a mold to both contain the resin and protect the sheath. The sheath is inserted through the mold, and then the cap is inserted, before pulling the assembly down into the occlusion mold. Resin is then added and left to cure prior to demolding.}
    \label{fig:mckibben}
    \vspace{-1em}
\end{figure}

The snake robot is actuated by McKibben pneumatic artificial muscles, which are arranged on the body to mimic aspects of two muscle groups that contribute to locomotion in biological snakes: the iliocostalis and the semispinalis. 
Our \textit{rib actuators} are derived from the iliocostalis, and attach to the vertebra and to the ribs, spanning across four vertebral segments and has a length of 115 mm.
Our \textit{body actuators} are derived from the semispinalis, and connect low on the vertebra to the protrusions at the top, which mimic the neural spine. These actuators also span across four vertebral segments, measuring 80 mm in length. 
The resulting design contains 7 body and 7 rib actuators on one side, with 28 actuators in total. 
Both types of actuators are designed to contribute to lateral bending of the robot. 

The fabrication procedure of McKibben actuators starts with inserting the PLA cap equipped with air tubing and assembly threads into the molded inner bladder (Ecoflex 00-30, Smooth-On). The bladder is inserted into the expandable PETG sheath, followed by the positioning of the occlusion mold (Ecoflex 00-30, Smooth-On) up to the PLA cap. Subsequently, two-part resin (Smooth-Cast 61D, Smooth-On) is poured into the cavity in the occlusion mold until it was filled. The primary role of the resin is to seal the cap interface and prevent air-leakage. Following resin curing, the same procedure was repeated for the other side of the actuator. Fabricated actuators were assembled to the robot as depicted in Figure \ref{fig:combinedSnakeDesignAndActuation}(d) by utilizing the threads attached to the PLA cap. While the sheath and tubing manufacturing occur in a similar way to most McKibben actuator manufacture, our capping procedure is designed to prevent failure or a dislodged cap during operation. The capping procedure is shown in Figure \ref{fig:mckibben}.
Note that the final curvature of the robot is predominantly determined by the McKibben actuators. While the vertebra can move up to 25$^\circ$ in either direction, the maximum actuator stroke and passive extension is reached much sooner. 
We relied on an identical activation pattern to maintain similar gaits in testing (discussed in Section \ref{sec:experiments}). 
McKibben actuators also provide structural compliance to the robot compared to the servomotors used in state-of-the-art robots that perform undulatory locomotion gaits \cite{wang_mechanical_2023}. However, this comes at a cost of lower performance consistency and actuation uniformity.
\subsection{Scales}

The scale structure is designed to mimic the ventral scales of snakes. Our scale structure design consists of three different units. 
The \emph{primary unit} of scales are directly attached to the ribs or to each other.
The approximate size of the scale structure is 10 mm in height, 24 mm in width, and 55 mm in depth, which means they are shorter in width than the vertebra.
This choice was intentional, as the scale structures are intended to be as independent of the robot body design as possible.
To match attachment points, the structures include two mechanical interlocks: they may either attach to the ribs and to adjacent scales, or just to the adjacent scales through a revolute joint. 
In our snake robot, approximately every other scale structure is connected to a rib. 
This design was inspired by a train track toy, and provides a robust and easy modular attachment while facilitating passive lateral bending. 
The \emph{secondary unit} of scale structure interlocks to the primary unit and defines the scale angle, ($\theta $). 
Hence, replacing the secondary unit is enough to adjust the scale angle. Secondary units measure 10 mm in height, 14 mm in width, and 52 mm in depth. 
The \emph{tertiary unit} of the scale structure is the scale itself which easily inserted into the groove in the secondary unit. They exhibit 0.5 mm of thickness.
Each component of the scale structure is shown in Figure \ref{fig:combinedSnakeDesignAndActuation}(b).
Finally, two elastomer cords were added to the design to incorporate passive elasticity.
These cords pass through holes in the scales, and have geometric interlocks to maintain the correct relative position. 
All components of the scale structures are 3D printed from PETG filament with an infill density of 15\%, and a layer height of 0.2 mm, while the elastomeric cords are molded from Ecoflex 00-30 rubber elastomer (Smooth-On). 
In total, the robot has 22 scale structures (Figure \ref{fig:combinedSnakeDesignAndActuation}).

\section{FRICTION EXPERIMENTS AND LOCOMOTION TESTS}
\label{sec:experiments}
\subsection{Friction Measurements}

\begin{figure}[!t]
\centering\includegraphics[width=1\columnwidth]{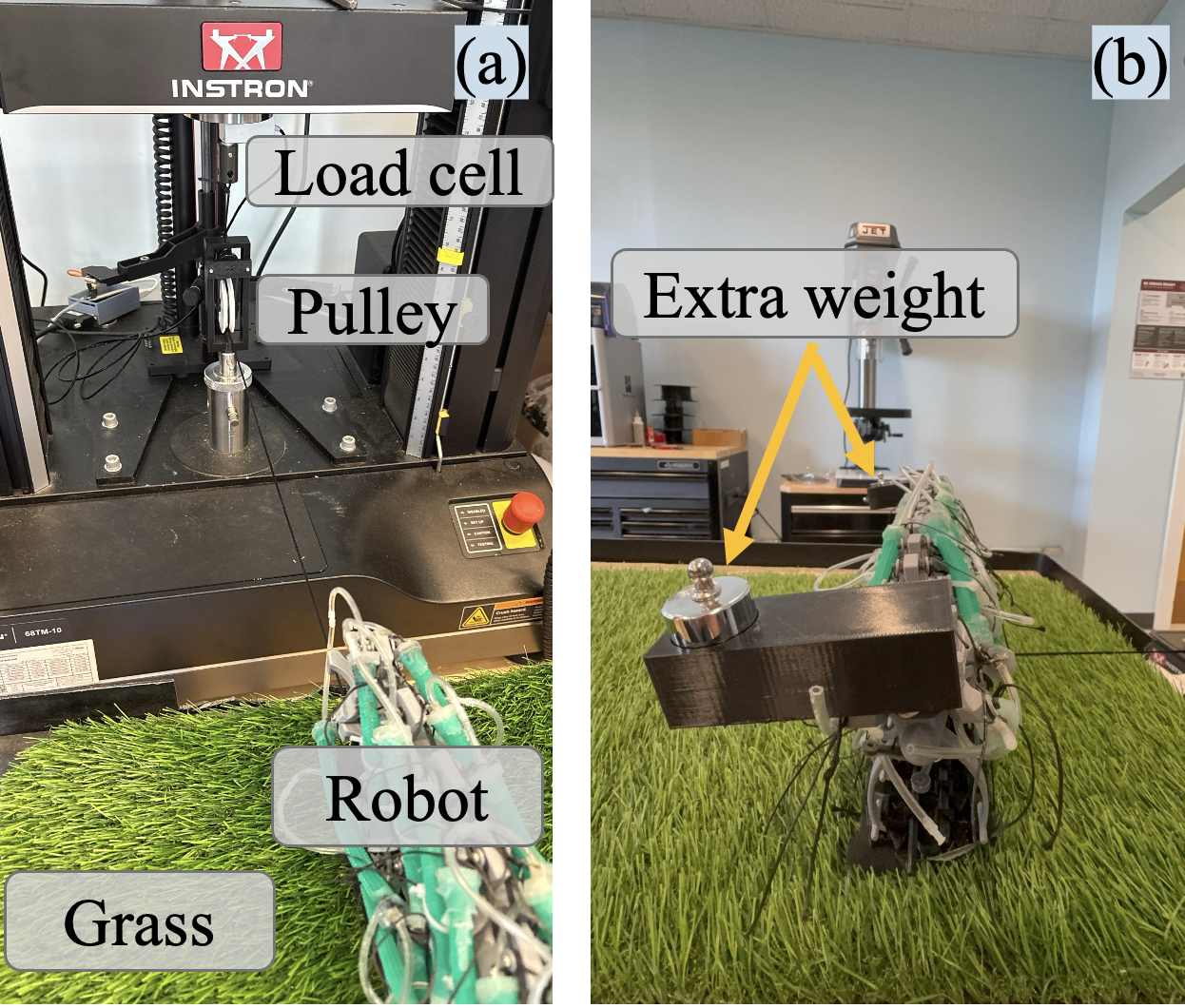}
    
    \caption{
     Experimental test setups for measuring (a) forward and backward, and (b) lateral friction. The displacement was transmitted to the robot via a pulley mechanism, with reaction forces recorded using a 100 N load cell.
    }
    \vspace{-1.5em}
    \label{fig:test_setup}
\end{figure}

We collected forward, backward, and lateral friction data with $\theta = 15^\circ, 25^\circ, 35^\circ, 45^\circ$ on bark, grass, carpet, and smooth surfaces. 
Grass and bark surfaces were selected to mimic the surfaces that snakes encounter in nature, to increase the practical real-world relevance of our work. The smooth surface was selected to evaluate the performance of our scales in the absence of any surface asperities. Finally, carpet surface, while more regular than most natural surfaces, was selected as a heterogeneous surface that is commonly encountered in indoor man-made environments (i.e., relevant to search and rescue applications). 
The measurements were performed with a universal testing machine (Instron, Norwood, MA) equipped with a 100N load cell and a custom test set-up. The test set-up connected a thread to the machine crosshead and ran it under a single pulley to change the direction to horizontal before connecting to the robot (see Figure \ref{fig:test_setup}(a)). 
The displacement limit of the universal testing machine is set to be 100 mm with a speed of 100 mm/min, resulting in 1 min per experiment. 
For each condition, the same trial was repeated 5 times. 
For example, forward-direction friction measurements with 25° scale angles on grass was repeated 5 times to verify the reproducibility of results. 
Directionality of the friction measurement was accomplished by changing the orientation of the robot. 
Lateral friction measurements required that the robot be oriented perpendicular to the pulling thread, but the combination of high friction force, low robot weight and tendency to catch either toppled or lifted the robot regardless of where the thread was attached (even if attached to the scales). 
We therefore added two weights, one either end of the robot, to create counter-torque and prevent tipping. 
This additional weight can be seen in Figure \ref{fig:test_setup}(b).

For each trial, we considered the final two-thirds of the time-series force data for subsequent analysis. Time-series force data were translated into forward, backward, and lateral friction coefficients with $\mu_f = F_f/F_n$, $\mu_b = F_b/F_n$, and $\mu_l = F_l/F_n$, respectively. 
The normal force $F_n$ denotes the weight of the robot, while $F_f$ is the frictional force measured by the testing machine's load cell. 
The robot has a mass of 636.25 grams for forward and backward friction experiments, and 867.47 grams for lateral friction experiments (due to the addition to the anti-tipping weights). 

\begin{figure*}[t!]
    \centering\includegraphics[width=1\textwidth]{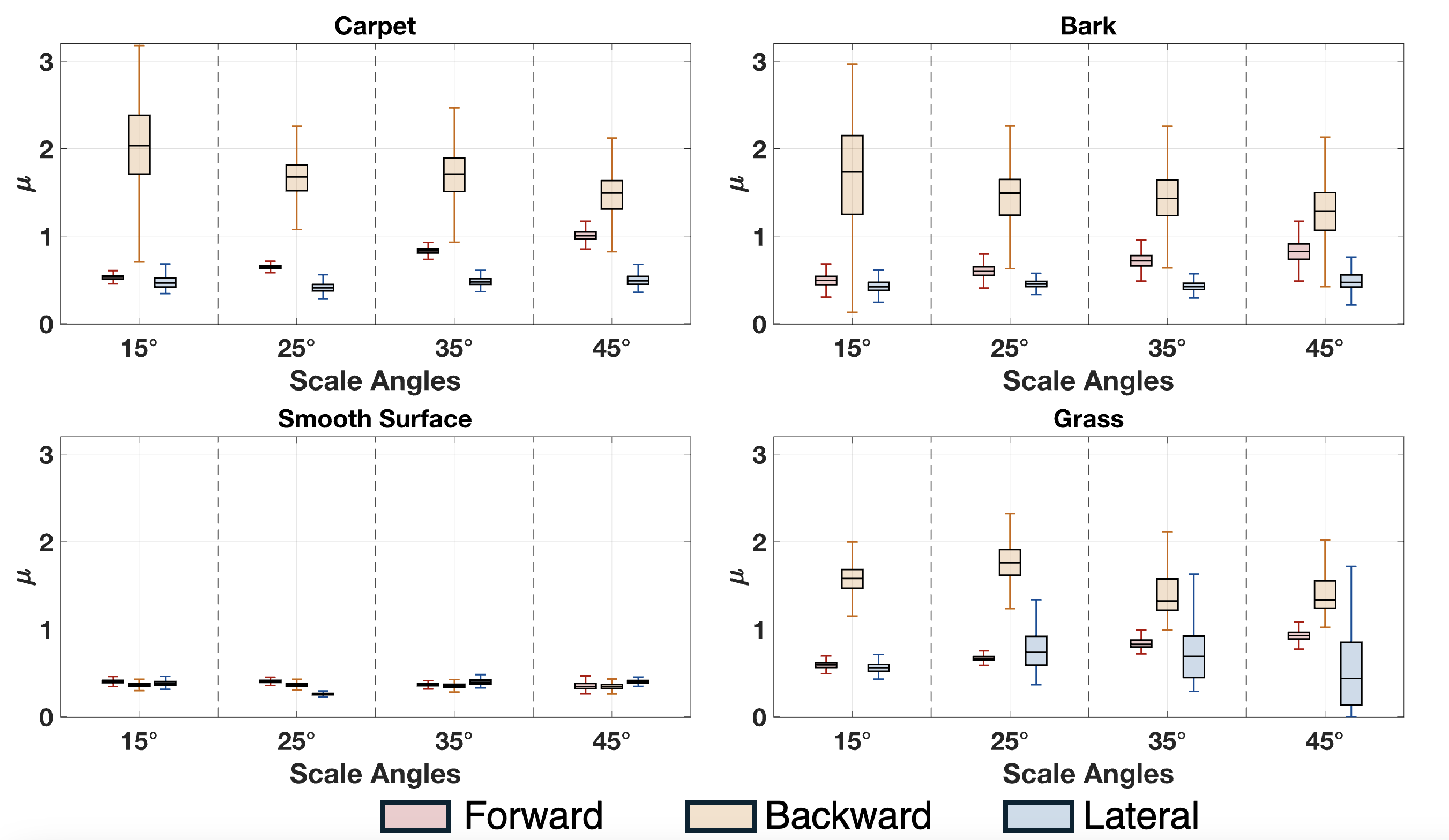}
    \caption{
    Boxplots illustrate $\mu_{f}$ (red), $\mu_{b}$ (orange), and $\mu_{l}$ (blue) distributions for each scale angle (15$^\circ$, 25$^\circ$, 35$^\circ$, 45$^\circ$) - surface (carpet, bark, smooth surface, grass) pair.
    }
    \label{fig:boxchartForRatios}
\end{figure*}

Figure \ref{fig:boxchartForRatios} summarizes the $\mu_f$, $\mu_b$, and $\mu_l$ friction coefficents across all scale angles and surfaces. 
Outliers that are beyond 1.5 times the interquartile range are excluded from the figure. 
This corresponds to a maximum of 4\% exclusion of the data, which was obtained in backward friction trials on carpet surface. 
This is followed by 3.86\% exclusion rate in lateral friction on carpet and 3.37\% in lateral friction on grass.
We attribute these outliers to transitory and atypical "catching" between the scale structure and features of the terrain, e.g., a single strand of false grass becoming lodged in the scale structure. 
Note that stick-slip behavior was observed in typical movement, similar to in existing work in the literature \cite{rafsanjani_kirigami_2018}, and the exclusion of outliers does not eliminate this observed stick-slip pattern from the data.  

Typically, backward direction measurements exhibit wider spread than forward direction trials across all surfaces and scale angles. 
This behavior can be attributed to the stick-slip pattern seen in backward direction trials in which the force builds up to the breakaway force, which is subsequently followed by a sudden decrease. 
In contrast, forward direction trials exhibit smaller variation around a dynamic friction force. 
Similar patterns for forward and backward frictional forces are also observed in works in the literature \cite{filippov_modelling_2016, rafsanjani_kirigami_2018}.

The magnitude of $\mu$ varies considerably with scale angle and surface. 
While comparatively higher friction observed on carpet, bark and grass surfaces, the maximum friction value remains near $\mu \approx 0.4$ on smooth surface. 
However, the maximum friction values on different surfaces are consistently observed in backward direction at $\theta = 15^\circ , 25^\circ$. 
The only exception is smooth surface in which $\mu$ values in every direction show little difference. 
This behavior was anticipated given smooth surface exhibits negligible level of surface asperities which is essential for creating anisotropic friction for snake locomotion. 
Predominantly, $\mu_l$ remains less than or comparable to $\mu_f$, which may account for the lateral movement we observed in locomotion experiments.

Anisotropic friction ratios illustrated in Figure \ref{fig:combinedSpeed_Velocity_Anisotropy} (bottom row) are obtained by dividing averages frictional coefficients. 
These averages are created by appending all data for a given set (e.g., carpet, 25$^\circ$, forward) and taking a time series average. 
The box-shaped marker correspond to anisotropic friction ratio $\mu_{b/f} = \mu_b/\mu_f$, while left-pointing triangle markers correspond to $\mu_{l/f} = \mu_l/\mu_f$. 
Red, green, blue, and black colors corresponds to 15°, 25°, 35°, 45° scale angles, respectively. 
Anisotropic friction ratios $\mu_{b/f}$ and $\mu_{l/f}$ exhibit similar decreasing pattern on carpet and bark, as $\theta$ increases. 
The correlation in grass varies at the lowest scale angle ($15^\circ$). 
The reason for this is unclear, but does not appear to be experimental variation, as the results for that scale angle-surface combination cluster as tightly as for other surfaces. 
On smooth surface, $\mu_{b/f}$ and $\mu_{l/f}$ ratios remains around 1 for all $\theta$, which shows that our scales are unable to generate anisotropic friction when surface asperities are absent, a phenomenon also observed in nature.

\subsection{Locomotion Experiments}

\begin{figure*}[t!]
    \centering\includegraphics[width=1\textwidth]{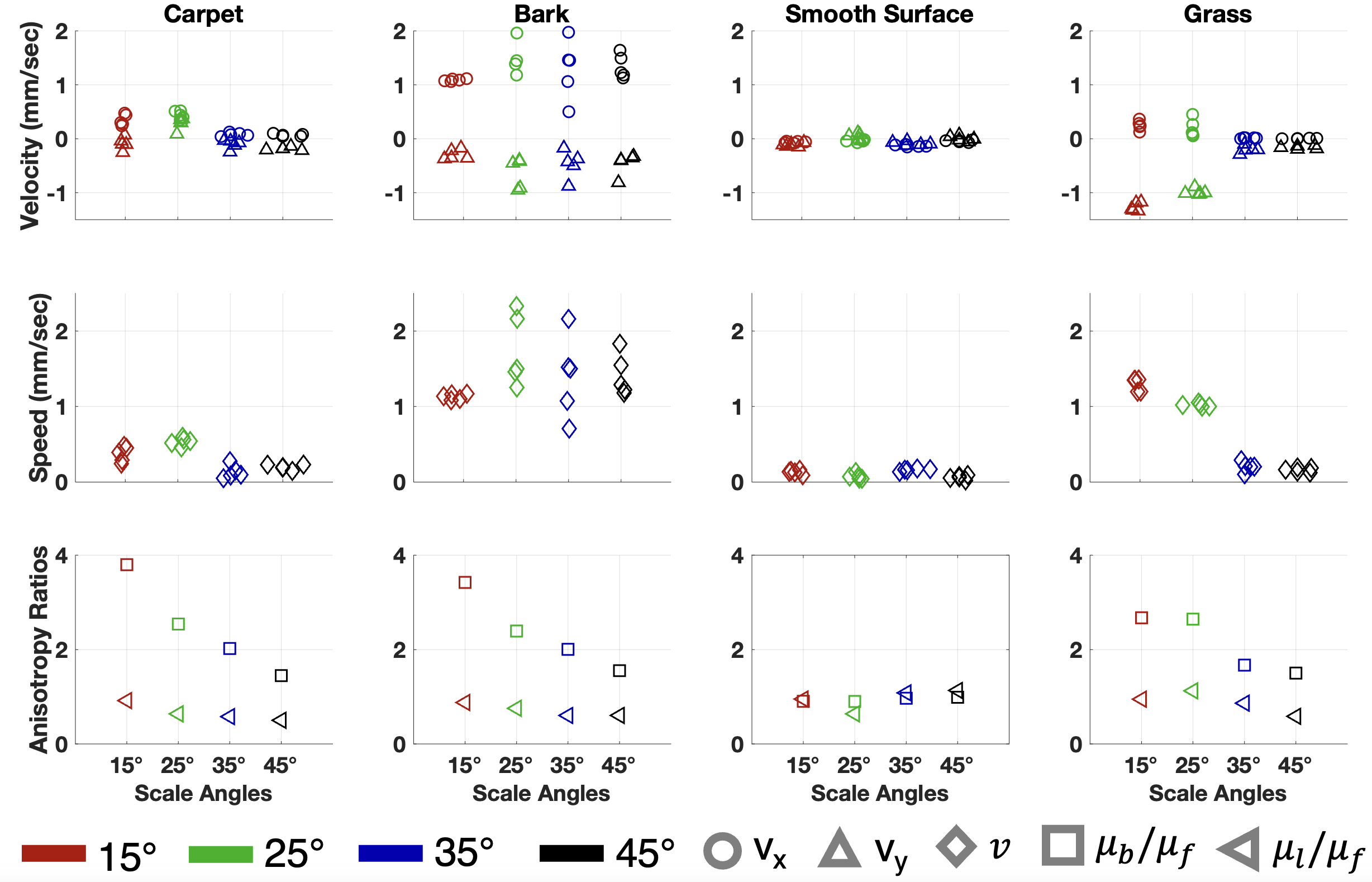}
    \caption{
    Velocity, speed, and friction ratio (top to bottom) values displayed for each scale angle-surface pair. Scale angles of 15°, 25°, 35°, and 45°  are depicted in red, green, blue, and black, respectively. The metrics $\mathbf{v}_x$, $\mathbf{v}_y$, and $v$ are denoted by circles, triangles, and diamonds, while friction ratios $\mu_{b/f}$ and $\mu_{l/f}$ are shown as squares and left-pointing triangles.}
    \label{fig:combinedSpeed_Velocity_Anisotropy}
\end{figure*}

For locomotion testing, the robot was initially located in the predefined location before each trial and McKibben actuators and scales were visually inspected to verify their assembly quality to the robot. 
A simple gait was created by cyclically activing the left and right side: all actuators on one side were inflated and deflated consecutively for 3.5 seconds with a controller system that consists of two solenoid valves  (Pneutronics) and a microcontroller (Elegoo Mega 2560), before switching to the opposite side. 
Each trial lasted 5 minutes and the displacement of the robot in x and y-direction were recorded at the end of the each trial. 
The only exception was bark surface in which each trial lasted 2.5 minutes due to robot leaving the bark surface in 5 minutes as a result of high speed.
The measurements were recorded with respect to the displacement of the bottom-left corner of the last scale. 
Each trial was also recorded with a smartphone (iPhone 16, Apple) for later-stage inspection. and analysis.

\begin{figure}[t!]
\centering\includegraphics[width=1\columnwidth]{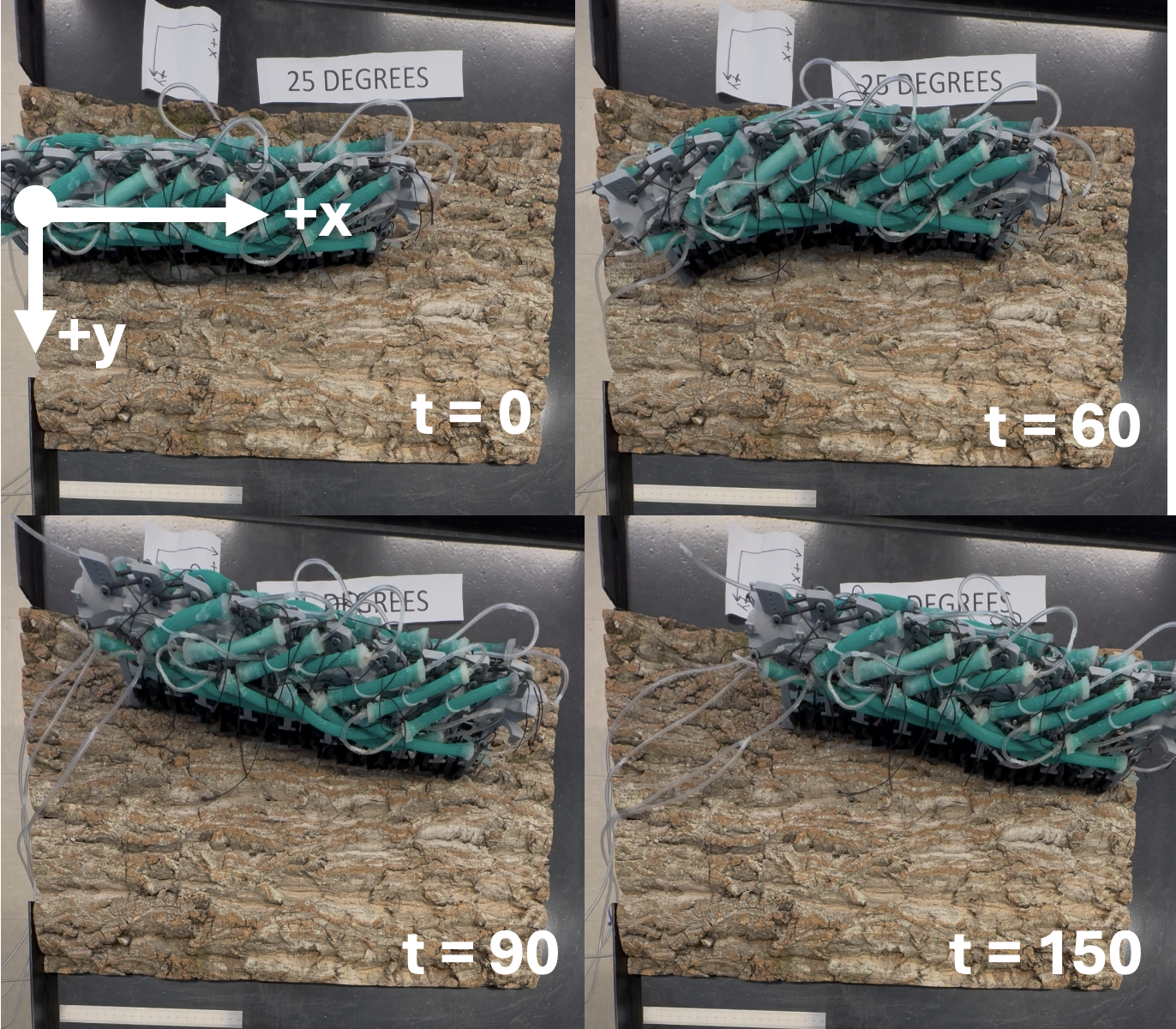}
    \caption{Snapshots of locomotion test on bark surface when $\theta = 25$° at \textit{ t} = 0,\textit{ t} = 60,\textit{ t} = 90, and \textit{t} = 150 seconds.}
    \label{fig:locomotionSnapshots}
    \vspace{-2em}
\end{figure}

In locomotion trials the robot is aligned with x-axis, hence, the forward motion is considered to be along the positive x-direction. 
Any motion along the y-direction is considered to be lateral motion, right side being positive and left side being negative with respect to the starting point as denoted in Figure \ref{fig:locomotionSnapshots}. 
Velocity performances in x and y direction are found as $\mathbf{v}_x= x_/t_f$ and $\mathbf{v}_y= y/t_f$ in which $x$, and $y$ denotes displacements in the x and y direction at the end of each trial, respectively. 
For grass, smooth, and carpet surfaces $t_f$ was taken as 300 seconds whereas it was 150 seconds for bark. 
Speed response for each trial was also evaluated with $v = \sqrt{\mathbf{v}_x^2+\mathbf{v}_y^2}$. Figure \ref{fig:combinedSpeed_Velocity_Anisotropy} displays velocity and speed responses for each surface - scale angle pair. 
Examples of locomotion trials are shared as a supplementary video.

\section{DISCUSSION}
\label{sec:discussion}

While friction coefficients were clearly and strongly related to scale angle for any given surface, and even showed similar trends across all surfaces expected to have asperities, we were not able to identify any general trends that tied friction ratios to locomotion speed across all surfaces and angles. 
In this section, we discuss our findings and posit an interpretation that suggests next steps.

\subsection{Smooth Surface}

Biological studies of snake locomotion have shown that snakes cannot locomote on surfaces with no asperities.
Note that this does not mean the surface has to have strong or even measurable frictional anisotropy. 
The surface simply must contain micro- or macro- features that interact with the frictional anisotropy created by snake scales \cite{hu_mechanics_2009}. 
While we do see a small amount of frictional anisotropy and locomotion when testing on a smooth surface, the displacements in both x and y directions are negligibly small (see Figure \ref{fig:combinedSpeed_Velocity_Anisotropy}).

\subsection{Grass, Carpet, and Bark}

While we are not aware of any prior works that try to correlate frictional ratios with locomotion speed on real or artificial grass, our results for grass do align with intuition developed from works for rectilinear gaits on other surfaces \cite{rafsanjani_kirigami_2018}. 
As the frictional ratios rose for lower scale angles, so did the locomotion speed (see Figure \ref{fig:boxchartForRatios}, final column).
For this surface, $|{\mathbf{v}_y }|$ was significantly higher than $|{\mathbf{v}_x }|$ when $\theta = 15^\circ, 25^\circ$, which means the movement was almost entirely sideways. 
With no control loop present, the movement direction is reflective of the \textit{resultant} force direction over the length of the body and gait cycle. 
Without consideration of gait, the $\mu_{l/f} \approx  1$ at $\theta = 15^\circ, 25^\circ$ implies that the cost of forward and lateral motion would be nearly the same, but every trial demonstrated lateral movement over forward. 
While the proposed theory for biological locomotion does consider gait, it retains a basic relationship between direction of lowest friction and and direction of movement \cite{hu_mechanics_2009}.
In our results, the movement preference is not readily explained by simple intuition, but may be impacted by complex scale-surface interactions, and the preference to move in the same lateral direction rather than equally likely to move left or right may be driven by subtle system imbalances.

On carpet, the robot did not exhibit any significant locomotion for $\theta = 35^\circ, 45^\circ$ and distinct - but slow - locomotory behavior appears only for $\theta = 15^\circ, 25^\circ$. 
In contrast, frictional ratios for this surface show a clear and distinct trend, with almost linear proportionality. 
Notably, $\mathbf{v}_x$ and $v$ at $\theta = 25^\circ$ is higher than $\mathbf{v}_x$ and $v$ at $\theta = 15^\circ$, although $\mu_{b/f}$ is smaller when $\theta = 25^\circ$.
The forward friction values for $\theta = 15^\circ$ had a particularly large spread, which is a possible cause for error, but does not fully explain the inconsistencies. 
While it would be reasonable to presume influence of friction based on weight of evidence from our work and others (higher $\mu_{b/f}$ values are present), the experimental results suggest that the path to predicting observed locomotion speeds is more complicated. 
Figure \ref{fig:combinedSpeed_Velocity_Anisotropy} displays that there is no significant inclination toward any y-direction on carpet surface. 

For bark surface, $\mu_{b/f}$ and $\mu_{l/f}$ values display a decreasing pattern with increasing $\theta$, as in carpet and grass. 
While $\mathbf{v}_x$ and $\mathbf{v}$ values does not appear to be directly influenced by $\mu_{b/f}$, evident rotational movement emerged for $\theta > 15^\circ$ in some trials, but not at $\theta > 0^\circ$. This can be seen in Figure \ref{fig:combinedSpeed_Velocity_Anisotropy} with $\mathbf{v}_y \approx$  -1 mm/sec at $\theta = 25^\circ, 35^\circ, 45^\circ$. This behavior might be associated with lower $\mu_{l/f}$, but evidence is inconclusive. However, this rotation may have contributed to the higher spread in speed values. Despite the rotation, the robot achieved the highest $v$, and $|\mathbf{v}_x|$ values on bark. 

In considering the results collectively, there is evidence of the influence of frictional coefficients - supported by existing literature - but comparing across surfaces identifies further inconsistencies. Frictional ratios were slightly higher in carpet than in grass, and yet locomotion speed in grass was faster and almost entirely lateral. Frictional ratios were similar between carpet and bark across scale angle, and yet locomotion was faster in bark. 
On grass and bark surfaces $\mathbf{v}_y$ tends to be predominantly negative, and on carpet, $\mathbf{v}_y$ exhibits a relatively more balanced distribution centered around zero compared to grass and bark.
In our initial evaluation of these results, we considered the potential for experimental error, but our results were broadly within expected variance for frictional measurements, and separate trials largely overlapped each other. Our experimental procedure also aligned with those used in similar publications \cite{lamping_frictional_2022,branyan_snake-inspired_2020}.

We therefore posit that, as surfaces become more complex, friction coefficients alone may not be sufficient to predict locomotion speed and direction.
Each of our test surfaces not only contains surface asperities of differing scales, but of differing aspect ratio, stiffness, arrangement, etc. 
On carpet, it was clearly observed that fibers did not have specific alignment for one direction, and appeared random. 
However, synthetic fibers on grass surface exhibited a clear alignment toward $-y$ direction. This surface characteristic could have introduced a bias to the underlying mechanics by making $-y$ the preferential direction for grass.
Similarly, bark was visually observed to be very likely to "catch" scales compared to the other surfaces, and this characteristic was not captured by either the average friction coefficient or the spread (carpet had a similar spread). 
This phenomenon is likely related to the stick-slip behavior noted in a prior publication \cite{rafsanjani_kirigami_2018}, and we posit that the importance of this idea will only increase as we attempt to study soft snake robots on 
heterogeneous surfaces.
As a path forward, we propose that a revised method of measuring friction is required, and that the test must mimic the desired dynamic movement of snake robots and distinguish between the interrelated ideas of friction and a tendency to catch on surface asperities.
Relying on one open-loop gait also limits the generality of the results, which we plan to address in the future with gait modulation.

\subsection{Biomimicry}

Marvi et al. \cite{marvi_scalybot_2011} reports that scale angles observed in that biological snakes are in the range of $20^\circ-30^\circ$. We observed ideal locomotion when $\theta = 15^\circ , 25^\circ$, and the highest speed was achieved at $\theta = 25^\circ$ on bark. Therefore, our modular pseudo-skin design yielded comparable preferred scale parameters as biological snakes. 
However, our tests showed undesired lateral motion and rotation. 
We posit that the mismatch between our scale's lateral friction and that of biological snake scales may be the cause. 
In literature, it has been shown that snake scales exhibit higher lateral friction than backward and forward friction \cite{hu_mechanics_2009}.
Conversely, we obtained lower friction values in lateral direction.

\section{CONCLUSION}
\label{sec:conclusion}


This paper presents an experimental friction and locomotion analysis for soft snake robot. We proposed an easily fabricated, novel snake scale design with adjustable scale angles. We conducted friction tests on grass, bark, carpet, and smooth surface to obtain forward and lateral anisotropic friction ratios for different scale angles, observing that both ratios decrease with increasing scale angle. 
We further attempted to correlate these ratios with locomotion performance on different surfaces. Results indicated that while higher locomotion speeds often came from higher anisotropic ratios on carpet and grass, a strong correlation or proportionality could not be established.
While this study serves as an initial step toward understanding soft robotic snake locomotion based on anisotropic friction, it is unable to capture the intricate mechanics behind the locomotion. In future, we aim to identify other contributing factors for surface-scale interplay.

\section*{ACKNOWLEDGMENTS}
\noindent We gratefully acknowledge the support of UMass Amherst, through the Faculty Research Grant and SPARC Grant.


\bibliographystyle{ieeetr}
\bibliography{ref}

\end{document}